\documentclass[conference]{IEEEtran}
\IEEEoverridecommandlockouts
\usepackage{cite}
\usepackage{amsmath,amssymb,amsfonts}
\usepackage{graphicx}
\usepackage{textcomp}
\usepackage{xcolor}
\usepackage{algorithm}
\usepackage{url}
\usepackage{todonotes}

\usepackage{algpseudocode}

\def\BibTeX{{\rm B\kern-.05em{\sc i\kern-.025em b}\kern-.08em
    T\kern-.1667em\lower.7ex\hbox{E}\kern-.125emX}}
\begin{document}


\title{Deduction Game Framework and Information Set Entropy Search}


\author{
\IEEEauthorblockN{Fandi Meng}
\IEEEauthorblockA{
\textit{Queen Mary University of London}\\
London, United Kingdom \\
f.meng@qmul.ac.uk}
\and
\IEEEauthorblockN{Simon Lucas}
\IEEEauthorblockA{
\textit{Queen Mary University of London}\\
London, United Kingdom \\
simon.lucas@qmul.ac.uk}

}

\IEEEoverridecommandlockouts
\IEEEpubid{\makebox[\columnwidth]{ 979-8-3503-5067-8/24/\$31.00~\copyright2024 IEEE \hfill} 
\hspace{\columnsep}\makebox[\columnwidth]{ }}
\IEEEpubidadjcol

\maketitle

\begin{abstract}
We present a game framework tailored for deduction games, enabling structured analysis from the perspective of Shannon  entropy variations. Additionally, we introduce a new forward search algorithm, Information Set Entropy Search (ISES), which effectively solves many single-player deduction games. The ISES algorithm, augmented with sampling techniques, allows agents to make decisions within controlled computational resources and time constraints. Experimental results on eight games within our framework demonstrate the significant superiority of our method over the Single Observer Information Set Monte Carlo Tree Search(SO-ISMCTS) algorithm under limited decision time constraints. The entropy variation of game states in our framework enables explainable decision-making, which can also be used to analyze the appeal of deduction games and provide insights for game designers.

\end{abstract}


\section{Introduction}

Deduction games represent an important category of games with a substantial user base engaging in regular play. Examples include single player games such as \textit{Black Box}\cite{black_box}, \textit{Mastermind}, and \textit{Wordle}, and multiplayer games such as \textit{Cluedo}, \textit{Scotland Yard}, and \textit{The Resistance}. While some poker games feature deduction elements, the term “deduction games” specifically denotes those where deductive reasoning is the fundamental mechanism\cite{games_journal_deduction}\cite{engelstein2022building}. The structure of single-player deduction games typically involves the concealment of a `secret code' at the game's start, prompting players to engage in a sequence of queries (actions). Following each player action, feedback is provided by the in-game `Oracle.' Players must continually engage in action and reasoning processes until they identify the secret.

Our longer term goal is to better understand single and multi-player
deduction games, and provide agents to play the games and tools to
analyse the games.  In this paper we focus on single-player games.  The main contributions are:

\begin{itemize}
    \item A framework for implementing single-player deduction games and different kinds of information sets.
    \item Information Set Entropy Search - ISES: a family of algorithms to play the games.
    \item Entropy variation as measure of a game's challenge or potential interest.
    \item Results on eight deduction games.
\end{itemize}

\section{Related Works}

Deduction games are a class of imperfect information games, in which \textit{Information Set} is a key concept: it
is the set of all game states that are consistent
with an agent's knowledge at a point in the game. The Information Set Monte Carlo tree search(ISMCTS) algorithm specializes in sampling potential states within an information set, effectively tackling the complexities and strategic subtleties inherent in games where complete information is not universally accessible \cite{cowling2012information}. It exhibits superior performance in deduction games, as demonstrated by Reinhardt's utilization of the SO-ISMCTS agent in \emph{Secret Hitler}, outperforming rule-based approaches \cite{reinhardt2020competing}.

Our work is closely related to Downing et al (2019) 
\cite{downing2019automatically}, who
used entropy considerations to derive 
a deduction game policy.  Although the core
principles of entropy-based policy are similar,
the methods are significantly different.  

Downing et al used a Domain-Specific Language (DSL) to 
describe deduction games.  

In our work we offer an alternative approach that has significant advantages:
\begin{itemize}
    \item No need to learn a DSL.
    \item Fast game solving in many cases, though this depends on the game and the variant of the algorithm used.
    \item The anytime variants of our algorithm offer extremely rapid game solving which often remains close to optimal. 
    \item We envisage that ISES can be more easily adapted to multi-player and/or impure deduction games i.e. games with a deduction element combined with other mechanics.
\end{itemize}

Besides, methods similar to ours have also been applied in research on specific games. For instance, Neuwirth and Kooi have used entropy-based methods in Mastermind\cite{neuwirth1982some}\cite{kooi2005yet}. However, our work focuses more on the versatility of the method and provides a deeper analysis of deduction games. Our final goal is to explore how entropy-based methods can be used to evaluate game states in information-based gameplay to optimize existing search algorithms. This application of entropy differs completely from some past studies that used entropy to optimize policy iteration\cite{xiao2019maximum}.

\section{Deduction Game Framework}

\subsection{Structure of Single-player Deduction Games}
\begin{figure*}[h]
\centering 
\includegraphics[width=0.62\textwidth]{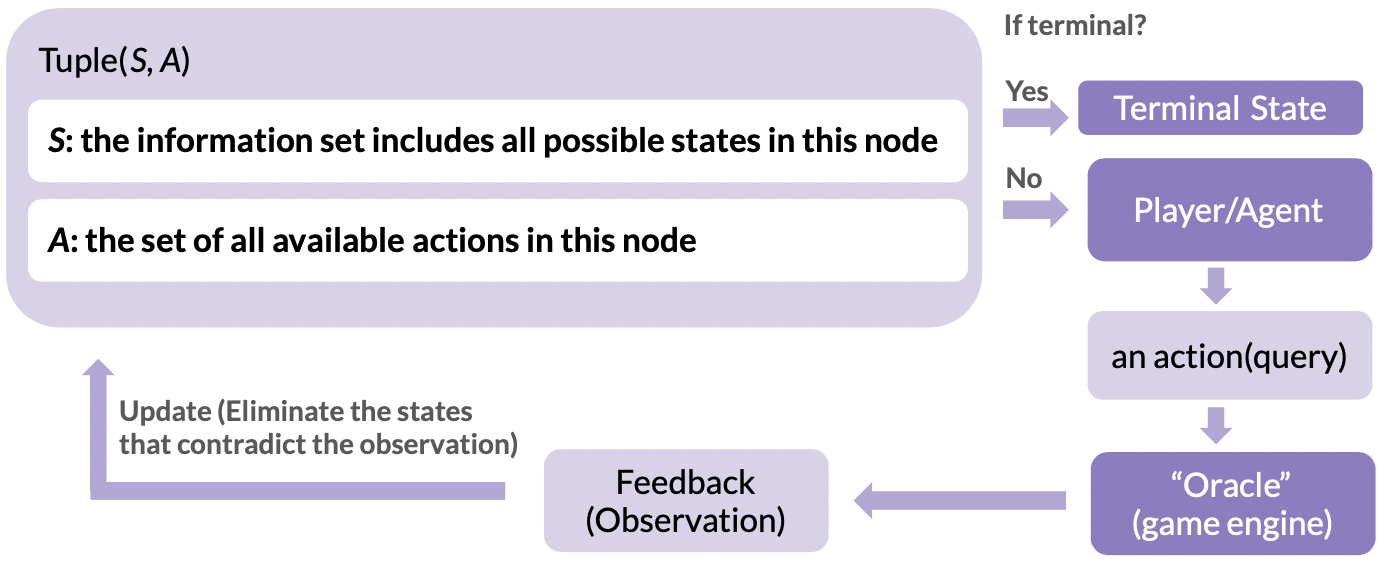}
\caption{The structure of typical single player deduction games}
\label{fig1}
\end{figure*}
Single-player deduction games usually start with a hidden 'secret', and players ask questions and receive feedback until they discover the truth. Unlike typical partially observable Markov processes represented by a 6-tuple $(S,A,T,R,\Omega,O)$\cite{kaelbling1998planning}, in these games, the 'Oracle' is usually truthful, and both feedback and state transitions are deterministic. Thus, transition probabilities can be ignored, and the reward form differs from other games since the goal is often to uncover the secret in the fewest steps possible.

A simpler model can be used to represent this process, as shown in Fig.~\ref{fig1}, where the game's decision-making information is represented by a 2-tuple. The information set is the collection of all possible states at a decision point, and the action space is all legal actions at that point. Without termination conditions, players must continually choose actions (make queries) and exclude possibilities contradicted by observations to update the information set. Termination conditions vary by game, usually ending when the information set has only one possibility, though different rules may apply to some games.

\subsection{Two representations of information sets}

In our framework, the information entropy of game information sets is used to evaluate the degree of uncertainty in an agent's incomplete information state within a game. Meanwhile, the goal of players in deduction games is to obtain certain state information, that is, to reduce the information entropy of the information set to a target value. We note that the information state of some games can be represented in table form, whereas for some games, due to the nature of combinatorial mathematics, their information states cannot be updated in a tabular form. The following two examples of games illustrate this point.

\begin{itemize}

\item \textbf{Fake Coin Game}: In this game, among n identical-looking coins, one coin is slightly lighter or heavier. Players can infer the identity and weight of the fake coin by placing coins on both sides of a balance scale and observing its tilt. The game terminates when the fake coin and its weight are determined\cite{smith1947counterfeit}.

\item \textbf{Mastermind}: In its classic form, the secret comprises a code of four colored pegs, with each peg chosen from a set of six colors. Players receive feedback by proposing their own four-color code. The number of black pegs in the response indicates the number of pegs with correct colors and positions, while white pegs represent correct colors but incorrect positions. The game terminates when the correct code is provided\cite{mastermind_wikipedia_2024}.

\end{itemize}

\begin{figure}
\centering 
\includegraphics[width=0.26\textwidth]{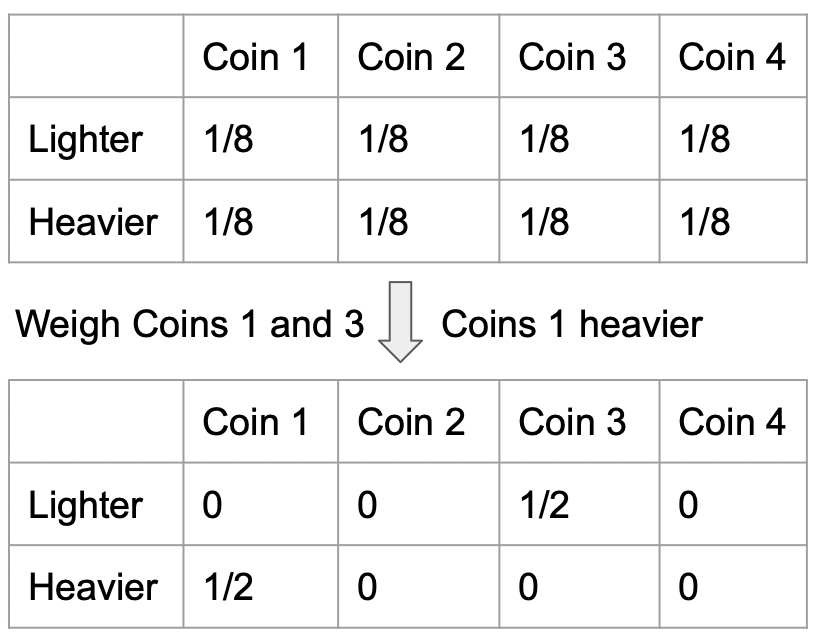}
\caption{The change in the information state matrix of the 4-coin fake coin game after a weighing result.}
\label{fakecoin}
\end{figure}

As shown in Fig.~\ref{fakecoin}, the information state of the fake coin game can be represented by a table, and the table can be accurately iterated at each step of the game to reflect the information currently held by the player. However, as an NP-complete problem, Mastermind cannot represent its information state as a two-dimensional matrix that can be iterated step-by-step due to its combinatorial nature\cite{berger2018query}. Consequently, obtaining its optimal solution cannot be transformed into a directly solvable polynomial computation problem\cite{stuckman2005mastermind}.

This leads to the need in our framework to calculate the entropy of information states in two ways. Based on Shannon's entropy formula (1) \cite{shannon1948mathematical}, for games like the fake coin game, it is only needed to calculate the entropy of the probability table of the information state. For games like Mastermind, however, it is necessary to know the number of all possible codes in the current information set. Since the probability of all possible codes is equal, the calculation formula is as shown in formula (2) and \textit{n} is the number of possible codes.

\begin{equation}
\begin{aligned}
H(X) &= -\sum_{i=1}^{n} P(x_i) \log P(x_i)
\end{aligned}
\end{equation}

\begin{equation}
H(X) = -\sum_{i=1}^{n} \frac{1}{n} \log \frac{1}{n}
= -n \cdot \frac{1}{n} \log \frac{1}{n}
= \log n
\end{equation}

\subsection{Entropy analysis of games}

\begin{figure}[H]
\centering 
\includegraphics[width=0.48\textwidth]{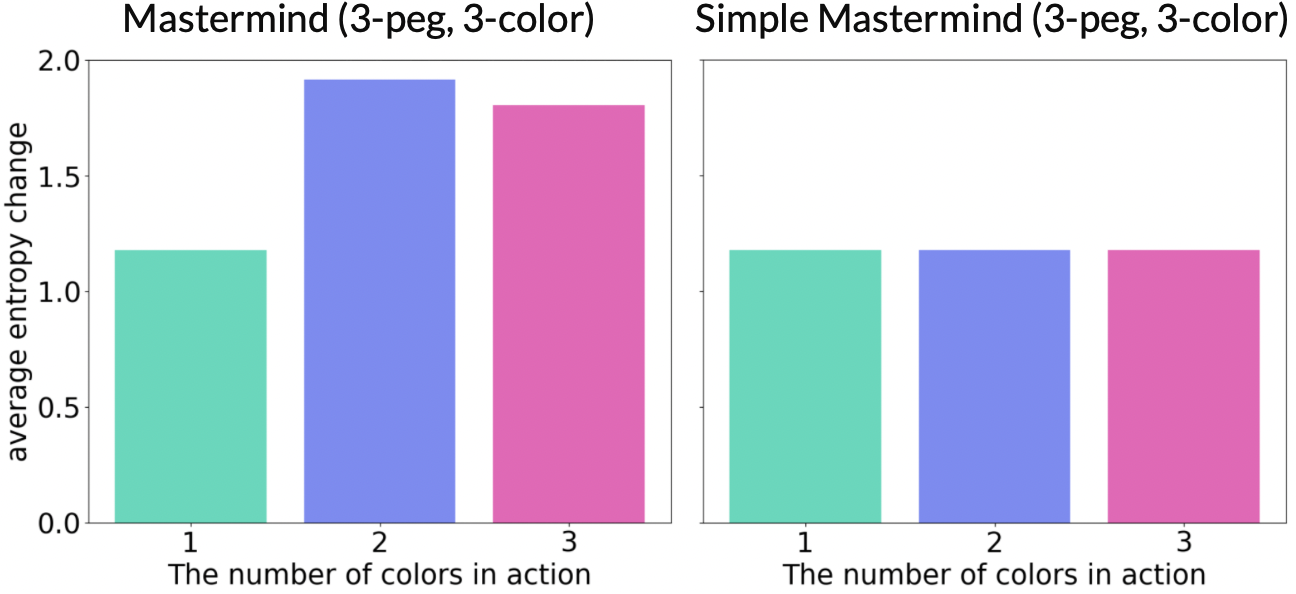}
\caption{The impact of different actions on the entropy change of game state information in the initial state.}
\label{fig2}
\end{figure}

The use of state information entropy to evaluate game states not only confers a high level of interpretability upon strategies but also manifests the strategic depth of a game. For example, in Fig.~\ref{fig2}, we demonstrate the mean value of the change in state information entropy induced by various actions at the initial state across all possible starting configurations in \textit{Mastermind} and \textit{Simple Mastermind}—games characterized by three pegs and colors. 

The bar graph for \textit{Mastermind} shows starting with two colors optimally reduces entropy and increases certainty. In \textit{Simple Mastermind}, unlike the standard version of \textit{Mastermind}, the feedback for guesses includes only black pegs, indicating correct colors in the correct positions, but no white pegs for correct colors in incorrect positions. Consequently, all first moves are similar, lacking strategic depth. Fig.~\ref{fig2} illustrates this difference, explaining why \textit{Mastermind} is more intriguing.

\begin{figure}[H]
\centering 
\includegraphics[width=0.36\textwidth]{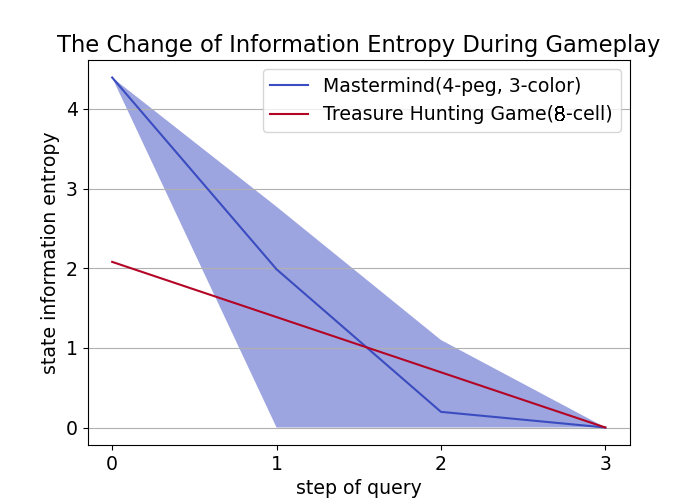}
\caption{The variation of state information entropy of two games over 20 experiments with the use of the optimal average entropy reduction strategy.}
\label{fig3}
\end{figure}

Is strategic superiority of some actions sufficient for an engaging deduction game? We visualized state information entropy changes through successive query rounds in \textit{Treasure Hunting Game} and \textit{Mastermind} Fig.~\ref{fig3}. Results assume an agent invariably adopting the optimal average strategy, with algorithmic details in the next section.

In \textit{Treasure Hunting Game}, players select cells to locate a hidden treasure. The game ends when treasure's location is narrowed to one cell. Fig.~\ref{fig3} shows entropy range over game iterations. For \textit{Treasure Hunting Game}, entropy change with optimal strategy is a straight line, while \textit{Mastermind} has a range. Binary search optimally locates treasure in 8 cells, consistently halving uncertainty. But \textit{Mastermind}, an NP-complete problem\cite{stuckman2005mastermind}, has uncertain outcomes even with the policy of minimizing entropy on average, lacking a polynomial solution like \textit{Treasure Hunt}.

This explains why \textit{Mastermind} is more interesting despite both being strategic games: \textit{Mastermind} has uncertain entropy changes and no definitive resolution, while \textit{Treasure Hunt} loses appeal once binary search is known. In conclusion, two factors make deduction games interesting: 1) Spatial - players' choices at decision points aren't equivalent. 2) Temporal - an optimal action's outcome remains uncertain after exploration, as \textit{Mastermind}'s entropy range in Fig.~\ref{fig3} shows. Same initial move could reduce entropy to zero or require more steps.

\section{Information Set Entropy Search}

In the previous section, we introduced a methodology for assessing game states using information entropy and discussed that the optimal action in single-player games should maximize average information gain. This idea has been expanded into a generalized search algorithm called the ISES algorithm.

Due to the exhaustive nature of the original ISES algorithm, which requires significant computational resources, we propose a variant that allows for controlled sampling scale. The pseudocode below shows that state and action space sampling scales can be manually set. By establishing a termination time, sampling can be halted upon reaching the predefined time limit, enabling immediate decision-making. This ensures suboptimal yet advantageous decisions within a manageable timeframe.

\begin{figure*}[h!]
\centering 
\includegraphics[width=0.8\textwidth]{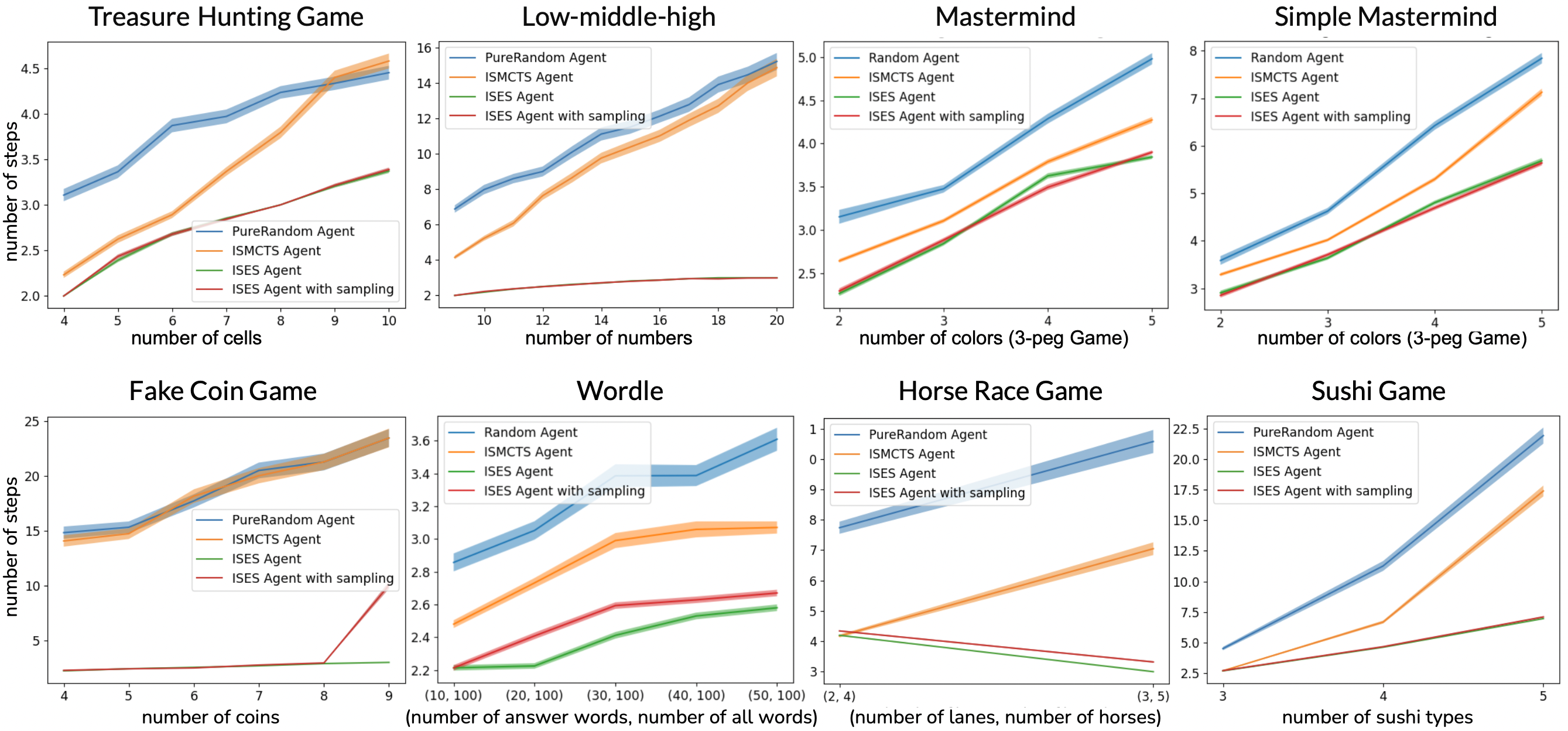}
\caption{The performance of four agents in different game sizes across 8 games.}
\label{fig4}
\end{figure*}
\begin{algorithm}
\caption{Sampling Information Set Entropy Search}
\begin{algorithmic}[1]
\State \textbf{Input:} Information set $S$, available actions $A$, sample size of states $m$, sample size of actions $n$
\State \textbf{Output:} An action $a$

\State Initialize $scored\_actions$ as an empty dictionary

\State $sampled\_states \leftarrow S$.randomsample($m$)
\State $sampled\_actions \leftarrow A$.randomsample($n$)
\State \textit{//By adjusting the sampling scale of the state space and setting termination time conditions for action sampling, the decision time can be limited to a fixed value.}
\For{each $action$ in $sampled\_actions$}
    \State $entropy \leftarrow 0.0$
    \For{each $state$ in $sampled\_states$}
        \State $observation \leftarrow state$.query($action$)
        \State $S'\leftarrow$ A deep copy of $S$ \State $S'$.update$(action, observation)$
        \State $entropy \leftarrow entropy + S'$.entropy()
    \EndFor
    \State $average\_entropy \leftarrow entropy/|sampled\_states|$
    \State $scored\_actions[action] \leftarrow average\_entropy$
\EndFor

\State $a \leftarrow$ action with minimum value in $scored\_actions$
\State \textbf{return} $a$
\end{algorithmic}
\end{algorithm}

\section{Results and Conclusions}

To assess the ISES algorithm's effectiveness, we tested it on eight games and compared it with two agents: the Random Agent and the SO-ISMCTS, which is effective in many imperfect information games. The goal of these games is to decipher the game's secret efficiently, so we designed the final reward to be inversely related to the number of steps taken.

For the eight games within our framework, we designed experiments across varying scales, with each of the four agents undergoing 500 trials for each variant. The results displayed in Fig.~\ref{fig4} encapsulate their average performance across these games. It is evident that the ISES algorithm consistently outperformed the others, achieving game completion in fewer average steps across all scales. 

The sampled version of ISES, even with a 0.1s decision time limit, closely mirrored the original ISES's efficacy, particularly in games with lower computational complexity. This indicates that for smaller-scale games, partial random sampling is sufficient to maintain performance because many actions and states are equivalent. However, as game complexity escalates, the sampled approach may falter, as seen in the \textit{Fake Coin Game} results, where exponential complexity growth at nine coins challenged the sampled ISES algorithm.

The experimental findings also highlight the limitations of the ISMCTS algorithm, particularly in games with extensive search spaces where performance diminishes due to insufficient simulation within the constrained decision time. In our studies, the ISMCTS agent, like the sampled ISES agent, was allotted a 0.1s decision time cap, yet its performance was significantly inferior. This gap widened with increasing game scale due to the exponential growth of the search space, as clearly shown in the \textit{Treasure Hunt Game} and \textit{Low-Middle-High}.

In future we aim to expand our framework to include multiplayer deduction games such as Cluedo. The dynamics of multiplayer deduction games involve maximising one's own
information while minimising the information
gain of our opponents, and entropy-based methods seem
a natural way to approach this.

\bibliographystyle{IEEEtran}
\bibliography{ref}

\end{document}